# Towards Next Generation of Pedestrian and Connected Vehicle In-the-loop Research: A Digital Twin Co-Simulation Framework

Zijin Wang, Ou Zheng*, Liangding Li, Mohamed Abdel-Aty, *Member, IEEE*, Carolina Cruz-Neira, *Fellow, IEEE*, and Zubayer Islam

*Abstract*—Digital Twin is an emerging technology that replicates real-world entities into a digital space. It has attracted increasing attention in the transportation field and many researchers are exploring its future applications in the development of Intelligent Transportation System (ITS) technologies. Connected vehicles (CVs) and pedestrians are among the major traffic participants in ITS. However, the usage of Digital Twin in research involving both CV and pedestrian remains largely unexplored. In this study, a Digital Twin framework for CV and pedestrian in-the-loop simulation is proposed. The proposed framework consists of the physical world, the digital world, and data transmission in between. The features for the entities (CV and pedestrian) that need digital twinning are divided into external state and internal state, and the attributes in each state are described. We also demonstrate a sample architecture under the proposed Digital Twin framework, which is based on Carla-Sumo Co-simulation and Cave automatic virtual environment (CAVE). A case study that investigates Vehicle-Pedestrian (V2P) warning system is conducted to validate the effectiveness of the presented architecture. The proposed framework is expected to provide guidance to the future Digital Twin research, and the architecture we build can serve as the testbed for further research and development of ITS applications on CV and pedestrians.

## I. Introduction

### A. Background

THE emergence of numerous Intelligent Transportation System (ITS) applications have made an extensive contribution to the transportation system in terms of safety, mobility, and energy consumption. Connected vehicle (CV), as a core component of ITS, has received vast attention since it was proposed in the late 20th century [1]–[3]. With the advancement in vehicle connectivity, computing power, and automotive control, the CVs are able to "talk" to other traffic actors (vehicle, infrastructure, pedestrian, etc.) and collaboratively make decisions and perform driving tasks.

Apart from the vehicles, pedestrian safety also raised much awareness, as it has the highest fatality rates among road accidents. Many ITS applications have been developed and investigated to enhance pedestrian safety in a CV environment. Due to safety concerns, most of the ITS technologies are developed and tested in a simulation environment, which is based on the assumption that the simulation environment is well-calibrated to represent real-world scenarios. However, this is not always true. Furthermore, there is no interaction between simulated entities and real-world traffic participants, causing a reduction in the validity of testing. Hence, an experimental environment that supports both CV and pedestrian in-the-loop is needed for the research and development of vehicle-to-pedestrian (V2P) ITS applications.

Digital Twin, as well as parallel driving, as an emerging technology in the transportation field, provides unprecedented opportunities to support the development of ITS applications and address the above issues [4], [5]. By definition, a Digital Twin is a digital replica of a physical entity in the real world [6]. In transportation, it could be the technology that projects all the traffic participants into a digital road network in a real-time manner. As real vehicle or pedestrian behaviors are reflected in a digital environment, the V2P applications can be implemented virtually. By such means, the effects of ITS applications in a dangerous scenario can be tested as there is no safety concern in the digital environment. Also, as the digital entities are projections of real traffic participants representing real-world traffic, the experiment results could be as convincing as field experiments.

In this work, we propose a novel Digital Twin framework to support ITS research on CV and pedestrians. A sample architecture under the framework is demonstrated, which incorporates a Sumo-Carla co-simulation platform for CV and a Cave automatic virtual environment (CAVE) for pedestrians. This research is the first to introduce the concept and technical detail of CV and pedestrian in-the-loop for a Digital Twin environment. The contribution of this study includes:
- Introduced a CV and pedestrian in-the-loop Digital Twin framework, which consists of physical world,

Zijin Wang, Ou Zheng, Mohamed Abdel-Aty, and Zubayer Islam are with the Department of Department of Civil, Environmental and Construction Engineering, University of Central Florida, USA (email: zijinwang@Knights.ucf.edu; ouzheng1993@Knights.ucf.edu; M.Aty@ucf.edu; zubayer_islam@knights.ucf.edu).

Liangding Li and Carolina Cruz-Neira are with the Department of Computer Science, University of Central Florida, USA (email: leoriohope@knights.ucf.edu; carolina@ucf.edu).

digital world, and the connection in between. The framework describes the features that need to be digitally twined for both CV and pedestrians, including their external and internal states.
- A sample architecture is presented to show a realization method of the Digital Twin framework. The architecture innovatively connects a CV simulation environment (Carla-Sumo co-simulation) and pedestrian simulator (CAVE). The functions of the sample architecture currently include bi-direction data transmission, pedestrian and driver state capturing, and V2P application display.
- Demonstrated a case study of V2P collision warning under occlusion condition that validates the proposed framework. The case study investigates the effect of V2P warning over baseline (no warning) and autonomous vehicles (equipped with AEB). The advantage of using the CV and pedestrian in-the-loop system is shown through the experiments.

The organization of the remaining paragraphs is as follows: Section II reviews the literatures of CV and pedestrian simulation as well as digital twin technology; Section III introduces the proposed Digital Twin framework; Section IV presents a sample architecture using Carla-Sumo co-simulation and CAVE; Section V is the case study and Section VI is a summary and conclusion.

## II. LITERATURE REVIEW

There are numerous research focused on CV technology, including vehicle connectivity, cyber security, driving assistance, automotive control, and cooperative driving, etc. Most of the studies were conducted in a simulation environment as field test is expensive and time-consuming. Microscopic traffic simulation is widely used to study CV applications' benefits on safety [7]–[9], mobility [10]–[12], and energy consumption [13]–[15]. Virtual simulators, powered by game engine, have become increasingly popular for CV research, as they can simulate environment perception, vehicle dynamics, and control algorithms [16]–[20]. Driving simulator is also an effective tool to test CV technology including safety warning system [21], [22], traveler information [23], and cooperative driving [18], [24], [25]. Recently, an increasing number of research adopted co-simulation technics for CV research. The co-simulation incorporates multiple simulation tools with different functionalities such as autonomous driving, traffic flow simulation, vehicular network and sensing technology [19], [26].

The research on applying intelligent transportation system applications to enhance pedestrian safety heavily relies on simulation. Various collision warning or avoidance systems have been proposed that powered by vehicle-to-pedestrian communication [27], [28], and tested in micro-simulation by modeling the vehicle and pedestrian behavior. The advent of virtual reality (VR) technology brings opportunities to create a high-fidelity environment to involve real pedestrians for research and testing. VR headset is the most adopted equipment to display a virtual environment because of reasonable price and comprehensive technical support. However, the motion sickness issue is still a major problem that affects user experience [29]. Recently, some researchers used the Cave automatic virtual environment (CAVE) for pedestrian research [30]–[32]. CAVE is a VR environment where projectors are directed to between three and six of the walls of a room-sized cube [33]. It has much less motion sickness compared to headsets and allows people to walk in the cube space freely.

Pedestrian motion, the way of simulating walking in the virtual environment, is another hard task for pedestrian-in-the-loop research. The traditional method to simulate pedestrian crossing behavior is to perform "shout test": participants shout their crossing decision when they intend to cross the street. Walking simulators such as VR treadmills are also used in some research [34]. When conducting experiments, the participant is wearing VR goggles and walking in-place on a bowl-like treadmill, while tied with a band on his/her waist to prevent falling down. It could simulate unlimited walking distance, but the walking experience is inconsistent and still quite different from real-world walking. Recent studies have tried walking in-house with a VR headset [35]. The issue with it is that the participant is easy to lose balance and prone to fall down. Locomotion, that pedestrian mark time, and use leg a band to detect pedestrian motion, is another method to simulate walking [34].

The above-mentioned efforts are CV or pedestrian research based on simulation in designed experiments. The emerging Digital Twin technology allows researchers to develop new ITS applications in a real-virtual hybrid environment [36]. There are already a few research studies that explored the applications of Digital Twin on connected vehicles and drivers. A Mobility Digital Twin framework is proposed in [37], which consists of physical and digital space of three components: human, vehicle, and traffic. A concept of Driver Digital Twin is introduced with the aim of bridging the gap between existing automated driving system and driver digitization [38]. Also, deep learning for security in Digital Twin of cooperative ITS is investigated [39]. Although the Digital Twin concept has been applied to vehicles, drivers and other traffic participants, and many simulation or field test efforts have been made for it, to the best of our knowledge, there is a lack of systematic framework and research tool that considers both CV and pedestrian in-the-loop for a Digital Twin environment. As they are among the two most important traffic participants on the road, it is necessary to build a Digital Twin architecture to facilitate the research and development of the next generation of ITS technology.

## III. DIGITAL TWIN FRAMEWORK

The proposed Digital Twin framework consists of the physical world, the digital world, and the data transmission between all the modules. The framework architecture is shown in Fig. 1. The elements inside the brown rectangle on the left are the digital replica of CV and pedestrians. The physical world is represented in the green rectangle on the right, in which the real drivers and pedestrians are involved. The blue rectangles are the attributes of CV and pedestrian that need to be digitally twined. The grey rectangle at the center contains the CV and pedestrian-related ITS technologies that will be applied.

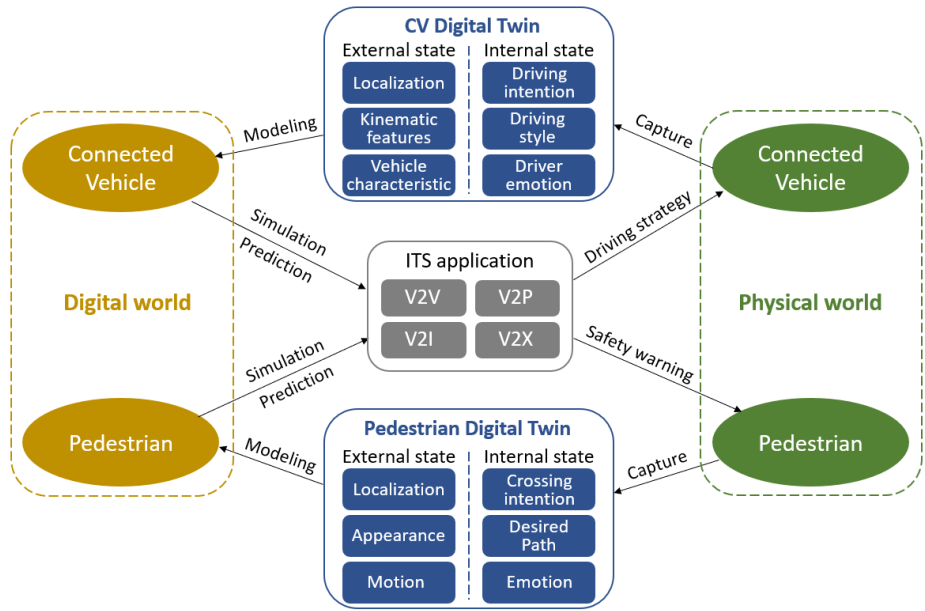

**Fig. 1.** Digital Twin framework for connected vehicle, pedestrian and traffic environment.

*A. Physical world*

The Physical world is the space where the real traffic participants (vehicle, driver, pedestrian, signal, and traffic sensors, etc.) interact with the environment. The challenge of digital twinning associated with the physical world is the state detection of the physical entities and the extraction of key features to send to the digital world. The traffic participants could be divided into living entities and non-living entities, and they may need different digital twining techniques.

For non-living entities, they only have external states, which are the physical states instead of psychological states, and their states can be fully captured and modeled through sensing or detection. For vehicles, the states include the kinematic features (e.g., position, speed, and acceleration) and ego characteristics (e.g., vehicle model and color). These features can be captured in real-time through onboard sensing (e.g., CAN BUS, GNSS) or external sensor detection (e.g., roadside cameras, radar). The digital twinning of traffic flow can also be challenging. The traditional method like Microwave Vehicle Detection Sensor (MVDS) provides reliable information about traffic flow in terms of volume, speed, and occupancy [40]. However, the MVDS data is still at a macro-level and the micro-level, or vehicle group level information is not captured. The increase in vehicle connectivity makes it easier for precise information collection of a vehicle group. In a mixed traffic environment, which may be a long-term state in the foreseeable future, the connected vehicles could build up a Vehicular ad hoc network (VANET) that collectively sense and share the information of the surrounding traffic.

The digital twinning for living entities is more complex than non-living entities, as both external state and internal state need to be detected or modeled. The internal state refers to the psychological status of a traffic participant including intention, emotion, personality, etc. The internal state normally cannot be measured directly and relies on identification or prediction through external behaviors. For drivers, their external state could be a set of human characteristics (e.g., gender, age) and wellness indicators (e.g., distraction, fatigue) that can be captured through wearable sensors (e.g., EEG sensor, EMG sensor), while the internal state includes driving intention and driving style preference. The internal state needs to be estimated and predicted using the external state as input, and various methods have been proposed including mapping attention field to external environment, RNN-based sequential models, and HMM models. For instance, the driver's gazing intention can be predicted through various methods [41]–[43]. For pedestrians, motion is the external state such as walking/running/standing and body gestures, and it could be measured by wearable sensing suites and devices, or through external sensing like video cameras. Pedestrians' internal state mainly includes crossing intention and desired path, and it also needs to be predicted [44], [45].

*B. Digital world*

The digital world is a virtual environment that parallels the real-world, where all the traffic participants are the projection of real entities in a real-time manner. Under the proposed Digital Twin framework, the digital world is the place where the vehicle and pedestrian-related ITS applications are deployed and simulated.

For connected vehicle digital twinning, the primary task is the projection of the vehicle states from the real world to the digital world. The vehicle's external state, including localization, kinematic features, and vehicle characteristics, are the features that could be directly replicated in the digital world. If considering the CV and drivers as an entity, the vehicle could thus have internal state due to the drivers' psychological attributes. The CVs share information about its ego states and the surrounding traffic, and thus they can conduct ITS applications such as safety warning, collaborative sensing, decision making, and vehicle control. Assisted by the computing power in the digital world, the effects of these ITS technologies can be simulated. Afterwards, the optimal driving strategy can be derived and feedback to the real-world CVs.

The digital pedestrian should also have the external state and internal state. The external state is relatively easy to be represented in the virtual space. It only requires the pedestrians' motion captured from the real-world to be digitally twined such as localization, moving speed and direction, and body gesture. The internal state, referring to pedestrian's psychological condition, needs to be modeled and predicted [46]. As the behavior of the digital avatar is the same as real-world pedestrian, their internal state (e.g., crossing intention, desired path) could be modeled and predicted through probabilistic models or machine learning methods [47].

Once the digital avatar of vehicles and pedestrians are replicated in the digital space, ITS applications could be implemented virtually. For a traffic scenario with both CV and pedestrian presence, various ITS technologies can be applied. Based on the communication subjects, it can be divided into vehicle-to-vehicle (V2V), vehicle-to-pedestrian (V2P), vehicle-to-infrastructure (V2I), or vehicle-to-everything (V2X) in general. In terms of functionality, these ITS technologies include driving cooperation, V2V or V2P collision warning, eco-driving guidance, etc. Normally, the digital world has stronger computing power (e.g., using cloud computing), and it can execute fast simulation and prediction to investigate the effects of the ITS technology and to generate optimal driving or crossing suggestions for CV and pedestrian, respectively. Afterwards, the suggestions are forwarded to the real world entities and form a data transmission closed loop.

## IV. Framework Realization

In this section, a sample architecture is presented to demonstrate a possible realization of the proposed Digital Twin framework, as shown in Fig. 2. The architecture is built based on various simulations and physical environments that incorporates both CV and pedestrians.

### A. Connected Vehicle Digital Twin

Since it is technically challenging and cost demanding to use a real vehicle for field tests, driving simulator could be a good alternative to represent a high-fidelity driving environment. Also, for experiments with safety concerns (e.g., a pedestrian crossing in an occlusion condition), simulator is the primary choice. In this sample architecture, a multi-driver co-simulation platform is adopted. The simulator is based on the co-

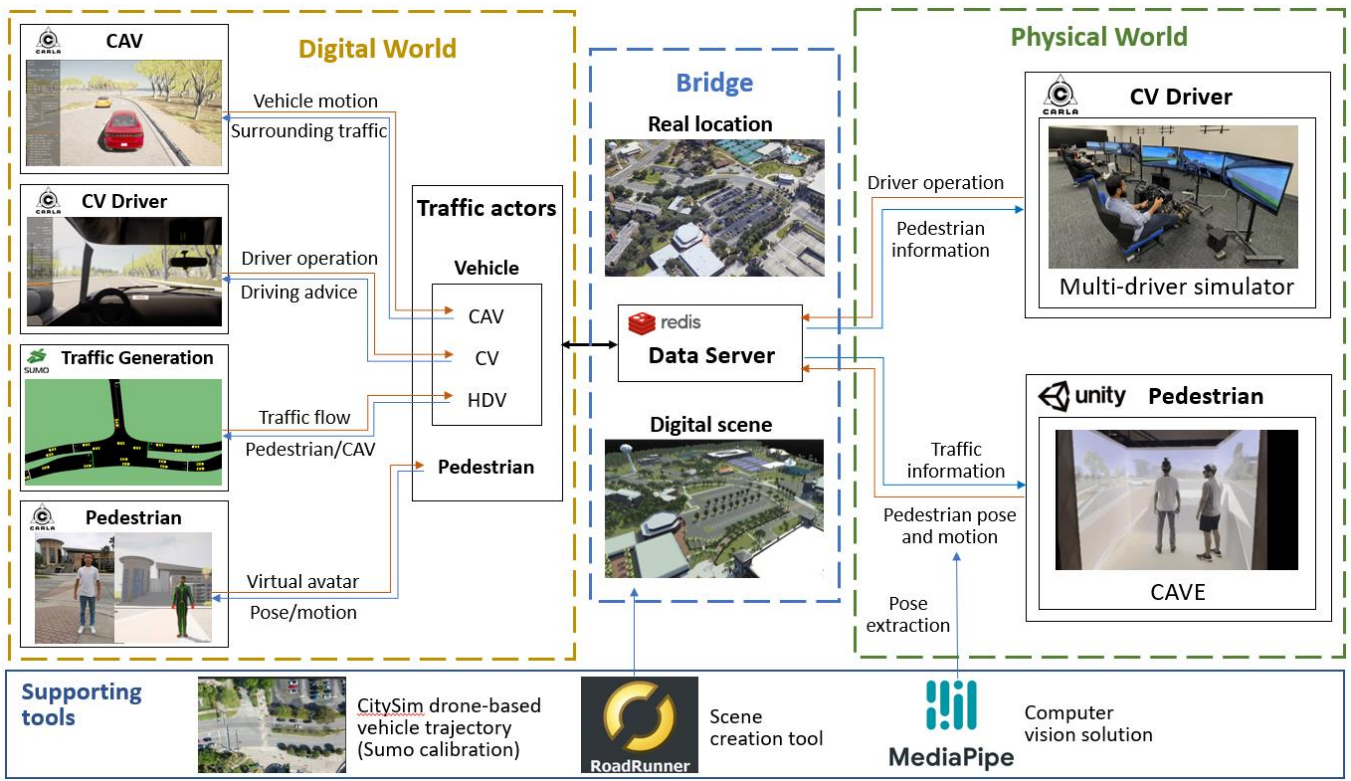

**Fig. 2.** Sample architecture of CV and pedestrian in-the-loop digital twin framework

simulation between Carla and Sumo, while multiple participants can drive simultaneously in the Carla virtual environment [48]. Fig. 3 demonstrates the Carla-Sumo co-simulation based on the location University of Central Florida (UCF) gym from CitySim dataset [49], which is an open-source vehicle trajectory dataset launched by our Safe&Smart Transportation (SST) team. Carla is an advanced autonomous driving simulator that is built on Unreal 4 game engine, and manual and autonomous driving can be conducted in Carla [16]. For driving simulator experiments, Carla is capable of rendering high-fidelity graphic displays and simulate vehicle dynamics. Sumo is an open-source microscopic traffic simulator [50]. It is powered by vehicle behavior models including driving behavior models and route choice models. After fine model calibration, Sumo can generate traffic flow that represents the real traffic and vehicle behaviors. For model calibration, vehicle trajectory data is adopted as it provides rich data on traffic flow and driving behaviors. The trajectory data used in this project is from the CitySim dataset. The Carla map and sumo road network are exactly matched as they share the same Opendrive road geometry file. The traffic actors are synchronized in both simulators, and Sumo vehicles and Carla vehicles can interact with each other to produce a mixed simulation environment. To power the co-simulation, a PC with Intel® Core™ i7-7800X CPU @ 3.50GHz × 12 and a memory of 64 GB plus NVIDIA GeForce RTX 2080 GPU is used. For the setup of driving simulator, we used triple 40'' Vizio TVs with 1920*1080 resolution, and Fanatec V2.5 steering wheel and hydraulic-supported pedals

To reproduce the real-world driving scene, high-fidelity 3D maps of the study location are built using the road scene designer RoadRunner (3D maps are available at CitySim dataset homepage: https://github.com/ozheng1993/UCF-SST-CitySim-Dataset). The maps are built using GIS data to guarantee the quality, including arial images, elevation data, and point cloud data. By such means, the driving behaviors of real drivers at real locations can thus be collected. Benefiting from using the simulator, the vehicle state can be directly accessed through Carla's Python API, instead of being detected or sensed as in real world. The drivers' state can be captured through detecting or sensing the same methods as real-world in-cabin technologies such as eye-tracking and fatigue identification. Once the driver and vehicle data are obtained, they are uploaded to the cloud server (Redis in the architecture) and stored. In addition, the connected and automated vehicle (CAV) and human-driven vehicle (HDV) are also included in the sample architecture in Fig. 2 to be able to simulate a mixed traffic environment.

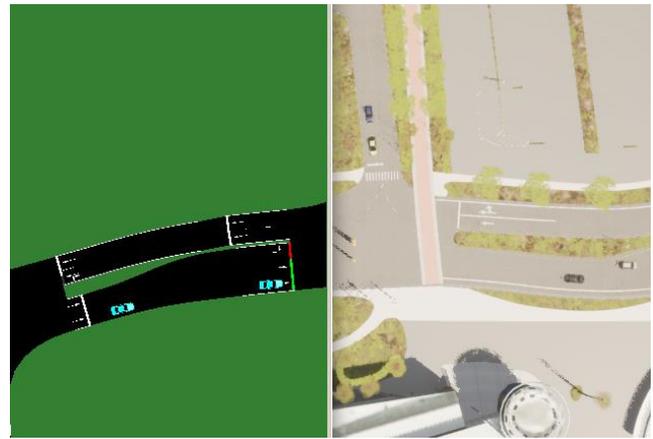

**Fig. 3.** Carla-Sumo Co-simulation (left: sumo; right: Carla)

*B. Pedestrian Digital Twin*

Safety is always the primary concern for pedestrian-related studies. When the pedestrian is exposed to potential risk in an experiment, it must be held in simulation instead of field testing. Due to this reason, the objective of pedestrian digital twinning is to create an immersive and close-to-real simulation environment for pedestrians.

In this architecture, we used a Cave automatic virtual environment (CAVE) to serve as the testbed for pedestrians. The CAVE equipment used in this study is called "VDen", which is a lightweight, portable version of CAVE. For this architecture, we used a PC with Intel® Core™ i7-10700 3.8GHz CPU, 32GB DDR4 RAM, NVIDIA GeForce 2080 GPU. The system projects 3D videos on 4 screens (left, front, right and bottom), and provides a 3*3-meter movable physical space. The projectors in this architecture are 4 BenQ ultra-short throw projectors support up to 1920x1200 120Hz 3D output, 60Hz per eye. The users are wearing a 3D glasses and a tracker for localization in the CAVE. Also, an additional tracker is used to track the eye position in order to get the correct perspective of view. Compared to the other VR device Head Mounted Displays (HMDs), CAVE has several evident advantages, especially for pedestrian simulation. First, it feels more natural and comfortable for the user only to wear a pair of lightweight 3D glasses and a tracker instead of a heavy headset. Second, CAVE brings less cybersickness because CAVE displays multiple images on the walls simultaneously instead of generating a new image when the user is moving his head. In addition, when experiencing VR in CAVE, users could naturally be aware of the physical presence of their own and each other's entire bodies, which is crucial for multi-user and social VR scenarios. Users can freely communicate with one another as if they were in the real world. In HMD setups, this is very challenging to achieve.

In order to replicate the pedestrian's pose in the digital world, the real body's keypoints are captured and matched to the digital avatar. Keypoint detection is carried out by Google MediaPipe Posekeypoint detector using pre-trained weights from BlazePose GHUM 3D. Unlike most state-of-the-art approaches, which depend on a robust server environment for

46inference, the MediaPipe achieves real-time pose detection on the 2022 MacBook pro with M1 Pro Max GPU. The MediaPipe outputs are uploaded to a Redis Pub/Sub server in JSON format, which includes 33 human key points' names and the bone transform in 3D space. This JSON file will be pushed into CARLA Client "WalkerBoneControl" class every tick to modify the CARLA walker agent's skeleton. To simulate the pedestrian's walking movement, locomotion method is used. It allows users to move forward or back by lifting their legs to simulate a real walking pose. The implementation is based on binding two additional lighthouse trackers to the user's legs and monitoring the range of movement of the sensors in the vertical direction to trigger a movement event. The horizontal orientation of the movement of the sensor also determines the direction of motion. The methods of pedestrian pose matching and achieving locomotion are shown in Fig. 4.

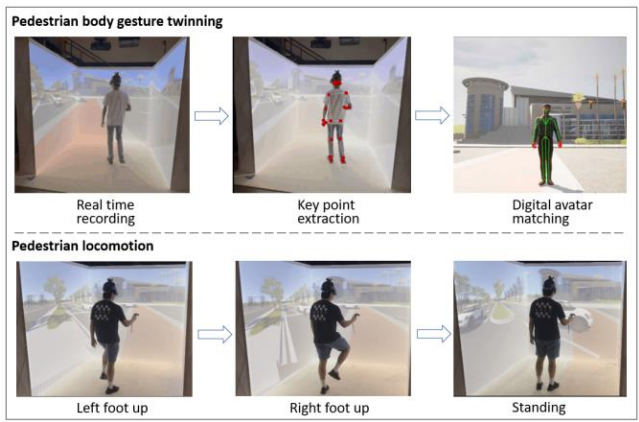

**Fig. 4.** Pedestrian Digital Twining in CAVE

*C. Closed-loop Data Transmission*

Since the presented architecture uses two major platforms: Carla-Sumo (vehicles) and Unity (pedestrian), the two digital spaces need to be synchronized. Fig. 5 shows the data transmission structure between the two platforms. The data transmitted from Carla-Sumo to Unity includes the vehicle state and driver state; while the data sent from Unity/CAVE to Carla is the pedestrian's location, motion and pose.

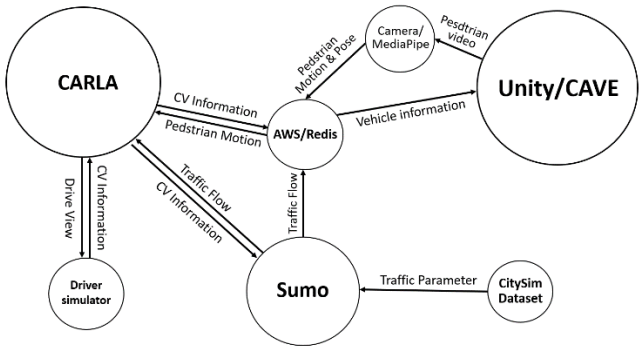

**Fig. 5.** Data transmission between each module

The idea of Digital Twin is to form a closed loop from physical entities to digital avatars and then generate feedback for physical entities. Hence, the proposed architecture follows this concept and keeps a closed-loop data transmission structure. When conducting an experiment, the driver and vehicle information is read from the driving simulator and is uploaded to Redis server. The pedestrians' location, motion and gestures are also tracked in the CAVE and are uploaded to the server simultaneously. At each simulation step, the system accesses the data server to download the vehicle or pedestrian data, and projects the avatar into the digital space. Afterwards, ITS technologies are implemented in the virtual world. For example, V2P collision warning system predicts crashing likelihood and triggers warning messages. Finally, suggestions for CV drivers or pedestrians are generated and feedback to the physical space (e.g., display warning messages on vehicle onboard unit).

## V. CASE STUDY

In this section, a case study based on the proposed architecture will be presented that investigates the effects of V2P collision warning system under occlusion conditions. Three experiments are designed to investigate V2P safety of (a) HDV without V2P warning; (b) AV with Automatic emergency braking (AEB); and (c) CV with V2P communication, respectively. In the experiment, HDV and CV are operated by the drivers on the driving simulator and AV is automatically controlled, while the pedestrian is simulated using the CAVE.

*A. Experiment Setup*

The experiment is derived from a real crash case in front of the gym of UCF (Fig. 3), where a pedestrian violated the traffic light and was hit by a vehicle under occlusion condition. During school hours, many students are heading to the gym and crossing behaviors with red-light violations are frequently observed, which sometimes leads to conflicts or even crashes. The development of multiple sensing and communication technologies allows the movement of both vehicles and pedestrian to be captured and broadcasted to each other, and V2P collision warning can be activated in a safety-critical scenario to prevent accidents. To test its effectiveness, three experiments are designed to reproduce the crash scene, as shown in Fig. 6. For all three experiments, the vehicle (HDV/CV driven by the driver, or CAV controlled by Sumo) is driving at 25 mph towards the intersection. When the vehicle's estimated arrival time to the conflict point is 5 seconds, the pedestrian light turns from "Don't walk" into "Walk", and the pedestrian in the CAVE starts to cross the street with a constant speed of 1 m/s [51]. Two buses are placed in front of the zebra crossing, and both driver and pedestrian cannot see each other until closer to the conflict point. For experiment 1, the driver and pedestrian will not receive any warning; in experiment 2, the AV activates the AEB system, and the vehicle and pedestrian are not receiving warning; while in experiment 3, the collision warning will be displayed through the smartphone (as on board unit OBU emulator) to both CV and pedestrian.

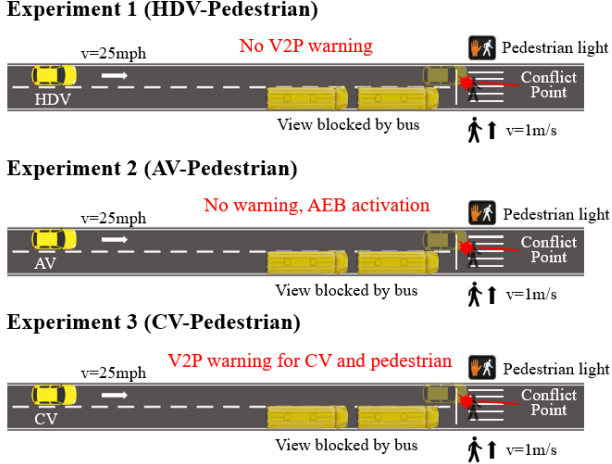

**Fig. 6.** Experiment design

The collision warning system consistently calculates the time-to-collision (TTC) between vehicle and pedestrian and triggers warning when the TTC is below the predefined threshold. The V2P warning triggering condition is shown in Equation 1.

$$Trigger = \begin{cases} yes, & if\ TTC < TTC^* \\ no, & else \end{cases}$$

$$TTC = |t_{veh} - t_{ped}| \quad (1)$$

$$t_i = d_i / v_i,\ i \in \{veh, ped\}$$

where $TTC^*$ is the activation threshold (1.5 second in this study), and $t_{veh}$, $t_{ped}$ are the estimated arrival time to the collision point, which is calculated by the remaining distance to the conflict point divided by current speed ($d/v$).

Once the system is triggered, a warning is sent to the CV driver and displayed through the Onboard Unit (OBU), which is emulated by a smartphone. For AV simulation, the P2V safety is ensured by activating the Automatic emergency braking (AEB) system. A Lidar is attached to the AV in Carla to detect the pedestrian, and once the V2P TTC is smaller than 1.5 seconds, the AEB triggers. This case study adopts the AEB algorithm of standard gradient deceleration on dry road [52], [53], and the braking profile is shown in equation 2.

$$Deceration\ (feet/s^2) = \begin{cases} 0;\ t < 0.25\ s \\ 65.7(t - 0.25);\ 0.25 \leq t < 0.6\ s \\ 23;\ t > 0.6\ s \end{cases} \quad (2)$$

where $t$ is the time to the AEB activation time.

Six groups (each group includes a driver and a pedestrian) of lab researchers participated in the experiments, and each group experience all three experiments. The participants are requested to behave in a safe manner and try to avoid potential crashes.

### B. Experiment Results

The vehicle decelerates once the pedestrian is detected, and the vehicle's distance to the zebra crossing when it completely stops (denote as V2P distance) can reflect the closeness of the V2P conflict. Braking point, which is defined as the vehicle's distance to the zebra crossing when the vehicle starts to brake, is used to measure the braking timing. A larger braking point indicates the driver starts to brake earlier to prevent a crash. Also, the average deceleration of the vehicle during the braking period are examined. The results of the average V2P distance, braking point and average deceleration for the three experiments are shown in Table I.

TABEL I
RESULTS OF V2P DISTANCE AND BRAKING POINT

| Experiment | V2P Distance (m) | Braking Point (m) | Average Deceleration (m/s²) |
|---|---|---|---|
| 1. HDV-Pedestrian | 3.941 | 11.458 | 7.464 |
| 2. AV-Pedestrian | 5.602 | 14.612 | 7.031 |
| 3. CV-Pedestrian | 17.179 | 28.325 | 6.216 |

As expected, the experiments of HDV-pedestrian show the smallest V2P distance and braking point, which means the driver starts to brake latest among all experiments and almost hit the pedestrian. The CV-pedestrian experiments have the largest value for both V2P distance and braking point. This indicates that the driver receives V2P warning before seeing the pedestrian and makes braking maneuver in advance to ensure safety. The value of V2P distance and braking point of AV-pedestrian is between the other two experiments. In the occlusion condition, the Lidar cannot detect the pedestrian until the occlusion is relieved, and the AEB activates to avoid a crash. The results show that under occlusion conditions, the HDV gets very close to the pedestrian without the V2P warning. For AV, the AEB system improves the safety slightly. With the collision warning for both CV and pedestrian, the V2P safety has been enhanced by a significant margin. In terms of average deceleration, the result of experiment 1 is the largest with a value of 7.464 $m/s^2$, indicating that the human drivers braked hard after seeing the pedestrian. The AV also generates an average deceleration over 7 $m/s^2$, while in experiment 3 the value is significantly smaller. Also, the average value of max deceleration among all the experiment groups are examined, and none of the braking behavior generates a max deceleration that exceeds 9.8 $m/s^2$ (8.812 $m/s^2$, 7.367 $m/s^2$, and 8.204 $m/s^2$ for HDV, AV, and CV, respectively). This shows that the V2P warning also enhance comfort while ensuring safety.

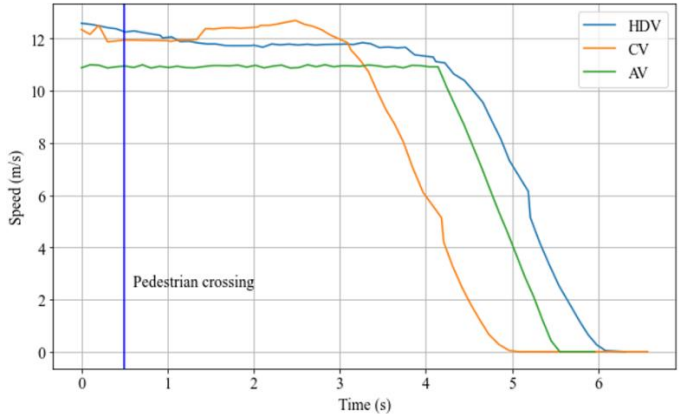

(a) Speed-time plot

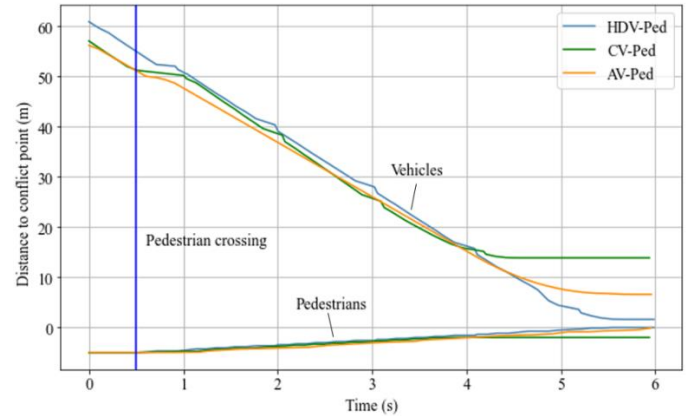

(b) Space-time plot

**Fig.7.** Visualization of results. (a) speed-time plot for three experiments; (b) space-time plot for both vehicles and pedestrians, note: the lines above x-axis are vehicles and the lines below are pedestrians

To further demonstrate the results, a group of experiments are visualized in Fig. 7. The speed-time plot is presented in Fig. 7 (a), where the speed over time of the HDV, AV and CV are represented in the blue, orange, and green line, respectively. It can be observed that the CV brakes earliest, followed by the AV then HDV. Fig.7 (b) shows the space-time plot of the three experiments. The space is the distance to the conflict point, and the upper lines belong to vehicles and the lines below the x-axis are pedestrians. The plot shows the CV stops far away from the conflict point while the HDV almost reaches the conflict point. In addition, the pedestrians in the HDV-pedestrian and AV-pedestrian group arrived at the conflict point, because they are not aware of the presence of vehicles. In the CV-pedestrian experiment, the pedestrian receives the warning and stops to cross the street, as the bottom green line stops to rise after 4 seconds. The differences in pedestrian behavior validate the need for both CV and pedestrian in-the-loop simulation, as conventional CV-related simulation study does not consider pedestrian reaction.

## VI. Conclusion

A Digital Twin framework that incorporates CV and pedestrian is proposed. The framework consists of the physical world, the digital world, and the data transmission between real-digital spaces. The attributes of the CV and pedestrian, including their external state and internal state are specified. A sample architecture of the framework is also built. The architecture combines two platforms: Carla-Sumo and Unity for the digital twinning of CV and pedestrian, respectively. The Carla-Sumo platform serves as the driving simulator to simulate a CV environment, while a CAVE powered by Unity is used to provide the simulation environment for the pedestrian. The data transmission between different platforms is defined and the method of forming a closed-loop data transmission structure is discussed. To validate the effectiveness of the proposed framework, a case study that investigates V2P collision warning system under occlusion condition is conducted based on the presented architecture. The results show that the V2P warning enhanced safety compared to HDV and AV. The case study also demonstrates the benefits of the proposed CV and pedestrian in-the-loop framework. The proposed Digital Twin framework is expected to serve as a powerful testbed for future research on ITS technology to enhance CV and pedestrian safety and mobility. The future work of this research is to expand the framework to incorporate other traffic participants such as cyclists, motorists and traffic light, and more complex traffic scenarios. Also, more complex scenarios like V2P safety application considering driver behavior and preference, vehicle connectivity, pedestrian motion prediction, and automated driving algorithms could be investigated using the proposed Digital Twin framework.

## Acknowledgment

The authors want to sincerely thank Dr. Amr Abdelraouf for his support of this research. The authors also want to thank Mr. Hanyang Liu at College of Environmental Design, UC Berkeley, for his help in architecture modeling and rendering of the 3D simulation map.

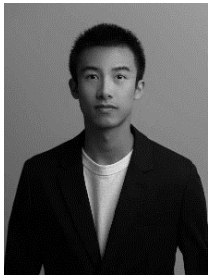

**Mr. Zijin Wang** is currently a PhD candidate at the Department of Civil, Environmental, and Construction Engineering, University of Central Florida, Orlando, FL, USA. He received the B.S. degree in logistics engineering from Central South University, Changsha, in 2020. His research interests include traffic safety analysis, intelligent transportation system, connected and automated vehicles, microscopic traffic simulation, co-simulation, and digital twin.

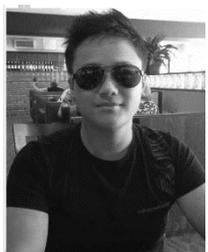

**Mr. Ou Zheng** is a computer vision engineer at the Civil Engineering department with the University of Central Florida. He has developed the UCF SST Automated Roadway Conflicts Identify System (A.R.C.I.S) and variety of vision-base research tools. His research interests include semantic segmentation, object detection, object tracking, 3d reconstruction, distributed computing and development of computer vision solutions.

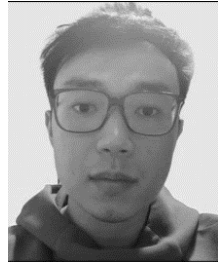

**Mr. Liangding Li** is a Computer Science Ph.D. student at the Computer Science department with University of Central Florida. He received his Bachelor of Science in the field of Electronic science from the University of Electronic Science and Technology of China (UESTC). His research interests include computer graphics, Virtual Reality (VR), 360 Video, VR for education, and human-computer interaction.

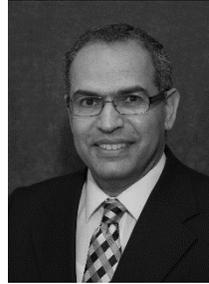

**Dr. Mohamed Abdel-Aty** (Member, IEEE) is currently a Pegasus Professor and the Chair of the Civil, Environmental, and Construction Engineering Department, University of Central Florida, Orlando, FL, USA. He has managed over 75 research projects. He has published more than 750 papers, more than 400 in journals (As of February 2023, Google Scholar citations: 27393, H-index: 89). His main expertise and interests are in the areas of ITS, simulation, CAV, and active traffic management. He received nine best paper awards from ASCE, TRB, and WCTR. He is the Editor-in-Chief Emeritus of Accident Analysis and Prevention

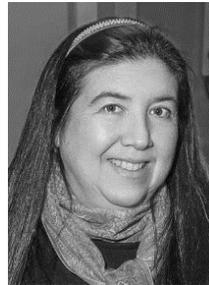

**Dr. Carolina Cruz-Neira** (Fellow, IEEE) is a member of the National Academy of Engineering. She is a Spanish-Venezuelan-American computer engineer, researcher, designer, educator, and a pioneer of virtual reality (VR). She is known for inventing the cave automatic virtual environment (CAVE). She previously worked at Iowa State University (ISU), University of Louisiana at Lafayette (UL Lafayette), University of Arkansas at Little Rock (UA Little Rock), and she is currently an Agere Chair Professor at University of Central Florida (UCF).

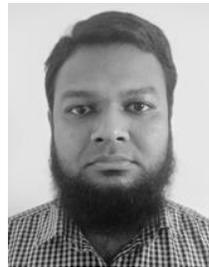

**Dr. Zubayer Islam** is a postdoctoral scholar at the University of Central Florida, USA. He completed his MS in Computer Science and PhD in Transportation Engineering University of Central Florida (UCF). He has worked extensively on developing an application for data collection, cloud computing, CV testing and used machine learning to estimate the vehicles' movements. He has also contributed to computer vision applications. He was also part of the team which won the USDOT safety visualization challenge. He has also been involved in developing Android mobile applications such as indoor and outdoor localization applications. His research interest lies with smart and safe transportation sensing, intelligent transportation system, machine learning and deep learning.